\pdfoutput=1

\documentclass[11pt]{article}

\usepackage[preprint]{acl}

\usepackage{times}
\usepackage{latexsym}
\usepackage{amsmath}
\usepackage{xcolor}
\usepackage{caption}
\usepackage{subcaption}
\usepackage{colortbl}

\usepackage[T1]{fontenc}

\usepackage[utf8]{inputenc}

\usepackage{microtype}
\usepackage{multirow}

\usepackage{inconsolata}

\usepackage{graphicx}

\title{User Profile with Large Language Models: Construction, Updating, and Benchmarking}

\author{
\textbf{Nusrat Jahan Prottasha}\textsuperscript{1}, 
\textbf{Md Kowsher}\textsuperscript{1}, 
\textbf{Hafijur Raman}\textsuperscript{1}, 
\textbf{Israt Jahan Anny}\textsuperscript{2} \\ 
\textbf{Prakash Bhat}\textsuperscript{3}, 
\textbf{Ivan Garibay}\textsuperscript{1},
\textbf{Ozlem Garibay}\textsuperscript{1} \\  
\textsuperscript{1}University of Central Florida, USA \\
\textsuperscript{2}Daffodil International University, Bangladesh, \textsuperscript{3}DotStar Inc., USA\\
Construction Dataset: \textcolor{red}{\href{https://huggingface.co/datasets/Nusrat1234/UserProfileConstruction}{\textcolor{red}{\texttt{https://huggingface.co/datasets/Nusrat1234/UserProfileConstruction}}}}\\
Updating Dataset: \textcolor{red}{\href{https://huggingface.co/datasets/Nusrat1234/UserProfileUpdate}{\textcolor{red}{\texttt{https://huggingface.co/datasets/Nusrat1234/UserProfileUpdate}}}}.
}

\begin{document}
\maketitle
\begin{abstract}

User profile modeling plays a key role in personalized systems, as it requires building accurate profiles and updating them with new information. In this paper, we present two high-quality open-source user profile datasets: one for profile construction and another for profile updating. These datasets offer a strong basis for evaluating user profile modeling techniques in dynamic settings. We also show a methodology that uses large language models (LLMs) to tackle both profile construction and updating. Our method uses a probabilistic framework to predict user profiles from input text, allowing for precise and context-aware profile generation. Our experiments demonstrate that models like Mistral-7b and Llama2-7b perform strongly in both tasks. LLMs improve the precision and recall of the generated profiles, and high evaluation scores confirm the effectiveness of our approach.

\end{abstract}
\section{Introduction}

\begin{figure*}[ht]
  \centering
  \includegraphics[width=1.00\linewidth]{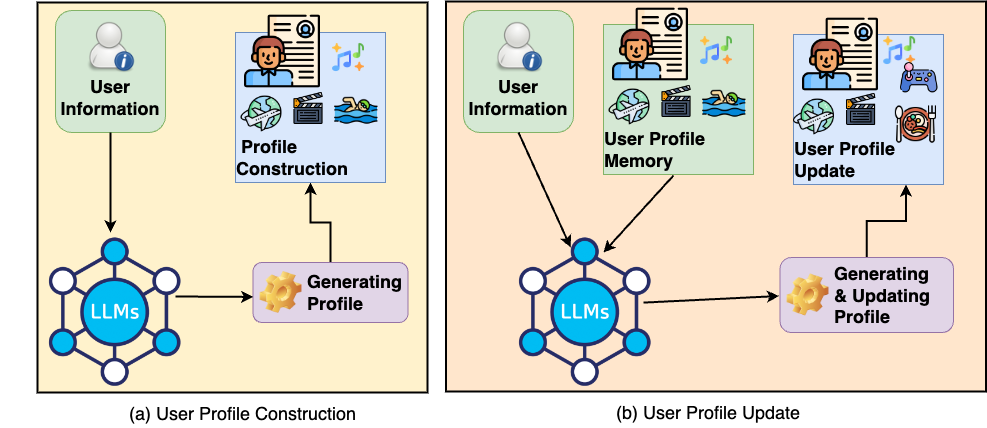}
  \caption{Overview of User Profile Management. Panel (a) shows profile construction from initial user data using an LLM, while panel (b) illustrates profile updating with new user information, maintaining dynamic User Profile Memory.}
  \label{fig:method}
\end{figure*}

User profiling is the process of constructing structured representations of individuals' preferences, behaviors, and attributes based on available data. In this study, we define a \textbf{user profile} as a collection of key-value pairs encapsulating information such as demographic details, interests, behaviors, and preferences. These profiles serve as the foundation for delivering personalized recommendations and user experiences across digital platforms.

In the digital era, understanding the complexities of user profiling on social media platforms is crucial. The pervasive use of user data for targeted advertising, personalized services, and social analytics has spurred the need for advanced, adaptable profiling strategies \cite{alt2009adaptive, carrara2011new, suh2005context, yu2012building, bartolomeo2008personalization, raad2010user, hazimeh2019reliable}. This task encompasses diverse factors, including users' occupations, educational backgrounds, and behavioral patterns \cite{ochirbat2018hybrid, preoctiuc2015analysis, chicaiza2015user}. Insights from these factors are vital for optimizing search algorithms \cite{agichtein2006learning}, enhancing friend recommendation systems \cite{xu2011using, ahmadian2019effective}, developing online marketing strategies \cite{saransomrurtai2011converting, dennis2016data}, and advancing computational social science \cite{zhang2020data, brandt2020identifying}.

Recent studies emphasize the growing complexity of user profiling due to the vast and diverse nature of user-generated content on social media \cite{luca2015user, ding2016predicting}. This complexity necessitates the integration of multiple data sources and advanced computational methods to effectively capture dynamic user behaviors \cite{alt2009adaptive, carrara2011new}.

Although regularly updating user profiles is crucial \cite{cameron2003digital}, there is a lack of open-source datasets and research focusing on efficient methods to reflect users' evolving preferences \cite{raad2010user}. Maintaining accurate profiles in a rapidly changing digital landscape is challenging, particularly for applications like natural language processing systems that rely heavily on user data \cite{hazimeh2019reliable}.

Social media platforms have become central to daily life, facilitating global communication and offering services from personal expression to community engagement \cite{zhang2020data}. These platforms support various activities such as networking, multimedia sharing, and content consumption, leading to an explosion of user-generated content \cite{brandt2020identifying}. This content provides rich data sources for user profiling research \cite{preoctiuc2015analysis}.

Early research treated user profiling as a multi-class classification problem, predicting characteristics like gender, age, and political orientation \cite{al2012homophily, kowsher2021gender,kowsher2020machine}. These models primarily used user-generated content and social interaction features \cite{suh2005context}. However, the field has progressed from basic demographic predictions to more nuanced analyses of user behaviors and preferences \cite{ochirbat2018hybrid}.

Recent advancements in deep learning have improved user profiling by providing deeper insights into user behavior \cite{ding2016predicting}. The field is moving beyond traditional approaches to explore patterns such as content consumption trends, sentiment dynamics in online interactions, and purchasing behavior changes over time \cite{saransomrurtai2011converting}. Emerging techniques like knowledge graph embeddings \cite{wang2017knowledge} and graph neural networks \cite{wu2020comprehensive} are increasingly used to integrate external knowledge and context, further refining user profiles \cite{godoy2005user}.

In this paper, we introduce two high-quality benchmark datasets specifically designed for user profile modeling: one for profile construction and another for profile updating. These datasets address a critical gap in current research by providing robust resources for evaluating user profiling techniques in dynamic scenarios.

Building on these datasets, we propose a novel approach that utilizes LLMs for both user profile construction and updating (Figure \ref{fig:method}). Our methodology employs a probabilistic framework to predict user profiles from input text, enabling precise and context-aware profile generation. This dual focus on static construction and dynamic updating reflects real-world needs, where user preferences and behaviors continuously evolve.

Our experiments demonstrate that models like Mistral-7b and Llama2-7b excel in both tasks. LLMs significantly enhance the precision and recall of generated profiles, and our evaluation metrics confirm the effectiveness of this approach. These findings highlight the potential of LLMs to advance user profiling by providing more accurate, adaptable, and context-sensitive models.

\noindent The primary contributions of this work are as follows:

\begin{itemize}
    \item We introduce two high-quality open-source benchmark datasets for user profile construction and updating, offering comprehensive resources for evaluating profiling techniques.
    \item We show a methodology leveraging LLMs to address both static construction and dynamic updating of user profiles, effectively capturing evolving user information.
    \item We conduct extensive evaluations of multiple LLMs, demonstrating their efficacy in user profiling tasks and providing insights into model performance.
    \item We present a dynamic profile updating mechanism that maintains the accuracy and relevance of user profiles over time, with significant implications for personalized systems and social analytics.
\end{itemize}

\section{Related Work}


User profiling is essential for personalized systems \cite{shen2005implicit, yao2020employing, zhu2008impact}, enabling platforms to deliver tailored recommendations \cite{balog2019transparent, lu2015exploiting, middleton2004ontological} and computational social media analysis \cite{arunachalam2013new, bamman2014gender, tang2015learning, al2020comparative}. Over time, profiling methods have evolved from rule-based systems to machine learning models, and now to LLMs, which offer improved accuracy and adaptability \cite{bloedorn1998using, wu2024understanding}. 

Early research treated user profiling as a \textit{classification problem}, focusing on predicting fixed attributes like gender \cite{liu2012using, rao2010classifying, liu2013s, sakaki2014twitter, priadana2020gender}, age \cite{rosenthal2011age, sap2014developing, chen2015comparative, fang2015relational, mac2017demographic, bessarab2025social}, and political views \cite{rao2010classifying, demszky2019analyzing, hettiachchi2021us}. used hierarchical classification to predict gender from Twitter data, leveraging linguistic features. \cite{liu2013using, ciot2013gender} expanded these methods to demographic and multilingual contexts, demonstrating the generalizability of text-based profiling. However, these models were limited by their static nature, as they couldn’t capture evolving user behaviors.

To address this, researchers began integrating social network data. \citet{al2012homophily} used \textit{homophily}—the tendency of people with similar traits to connect—to predict political affiliations. \citet{onikoyi2023gender} combined social interactions with text features to enhance gender prediction accuracy. Despite these advancements, these approaches still struggled to adapt to changes in user preferences over time.

To improve profiling, hybrid model emerged, combing textual, social and behavioral data. \citet{agichtein2006improving} showed that clickstream data and user interactions could enhance search algorithms. \citet{agarwal2013collaborative} and \citet{priadana2020gender} improved friend recommendation systems by integrating both content and network features. In marketing, \citet{bucklin2003model} and \citet{shah2021marketing} used browsing and purchasing behaviors to target ads effectively.

While dynamic profiling addressed many challenges, the introductions of LLMs revolutionized the field. LLMs can generate context-aware, nuanced profiles and adapt to evolving user behaviors. \citet{wu2024understanding} analyzed the role of user profiles in personalizing large language models, revealing that historical personalized responses are key to effective personalization and that profile placement at the beginning of the input context has a greater impact.

While significant progress has been made in developing model based techniques in user profiling, such as deep learning and reinforcement learning approaches, there is a notable lack of open source datasets to support comprehensive evaluation and benchmarking. This gaps limits reproducibility and hinders further advancements in the field. To address this, we propose tow novel open source datasets: one for user profile construction, providing diverse user information for generating accurate profile, and another for profile updating, capturing temporal changes in user behavior to evaluate dynamic profiling models. These datasets aim to facilitate transparent, reproducible research and drive innovation in user profiling.

\begin{figure}[ht]
  \centering
  \includegraphics[width=\linewidth]{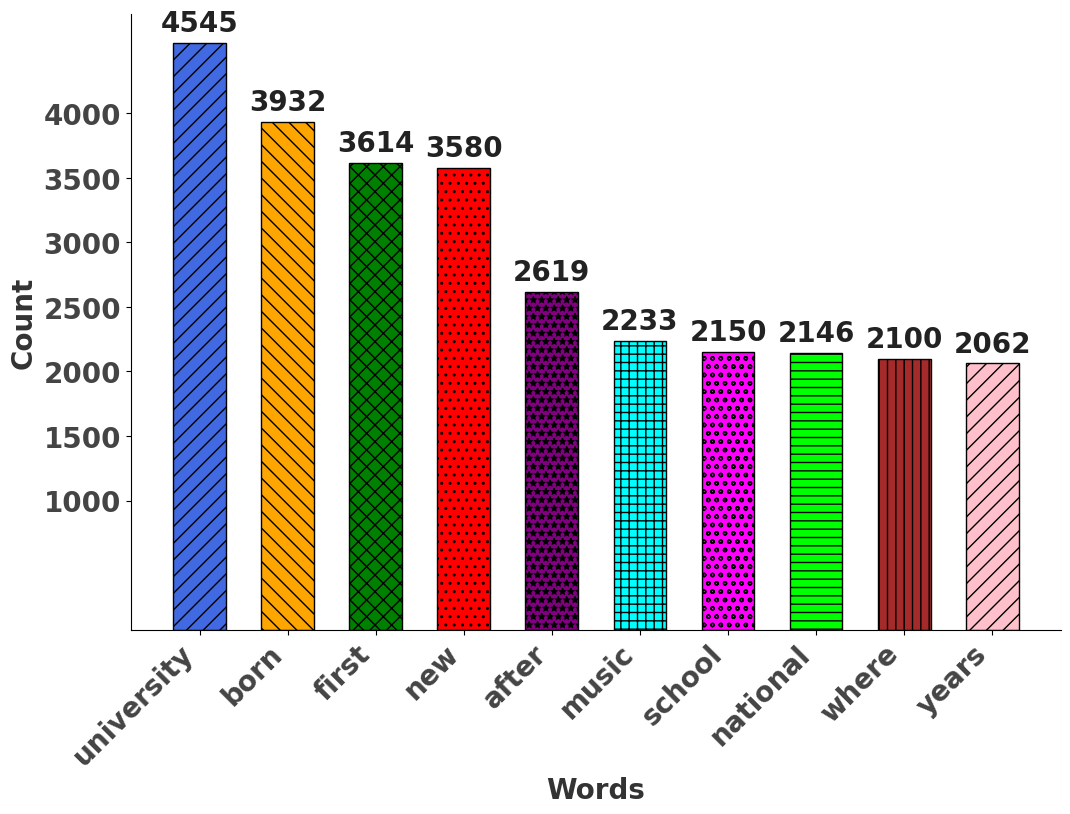}
  \caption{Top 10 Most Frequent Words. This bar chart illustrates the ten most frequently occurring words in the dataset, highlighting key terms such as "university," "born," and "first." The word "university" appears most frequently (4,545 times), followed by "born" (3,932) and "first" (3,614). The distribution suggests a strong focus on biographical and educational information within the user profiles.}
  \label{fig:bar_org}
\end{figure}

\section{Dataset Construction}

We built a dataset of user profiles from textual descriptions using the Wiki People dataset \cite{mahajan2020people}. 
This dataset has 42{,}786 entries. Each entry includes a Uniform Resource Identifier (URI), a person's name, and a detailed biography (details in Appendix Table~\ref{tab:overloaded_summary}). 
On average, each biography is about 1{,}896 characters long. This provides a lot of information about each individual's life, achievements, and important events. 
Such detailed data is useful for many NLP tasks, including sentiment analysis and entity recognition.  Figure~\ref{fig:bar_org} shows the most 10 frequent words of wiki people.

A notable feature of this dataset is the wide range of attributes—17{,}189 in total—which cover basic personal details and more in-depth characteristics. 
These attributes make the dataset valuable for a variety of analytical methods. Additionally, the use of URIs allows researchers to easily find more information if needed.

\subsection{Construction User Profile Dataset}
To construct our user profiles, we used Language Model APIs, such as Gemini \cite{team2023gemini}. 
We designed prompts that took the Wiki People data as input to generate user profiles. 
For example, when we provided information about ``Timothy Grucza'' and a short prompt, the API created a user profile for him. 
Repeating this process for all entries in Wiki People gave us 42{,}786 user profiles.

Below is a simplified example of the prompt and the generated output:

\noindent
\textbf{Input:}  Timothy Gruzca (Wiki data)

\textbf{Prompt:}
\begin{quote}
``\#\# Instruction: Given a profile information: \{data\}, generate a user profile with details like name, birthplace, profession, hobbies, likes, dislikes, and anything else in the data. If something is not mentioned, do not add new details. Use this format:

\#\#Example Output:

\textbf{Name:} Albert Einstein

\textbf{Likes:} History; Science; Traveling

\textbf{Education:} PhD in Physics from the University of Zurich

Do not include any extra text outside the user profile. Keep it short.''
\end{quote}

\textbf{Output:}
\begin{quote}
\textbf{Name:} Timothy Gruzca

\textbf{Born:} 1 July 1976, Melbourne, Australia

\textbf{Profession:} Cameraman, Documentary Film Maker

\textbf{Based In:} Paris, France

\textbf{Likes:} Working in conflict zones; Telling stories of people affected by war; Documenting humanitarian crises

\textbf{Dislikes:} War; Injustice; Suffering

\textbf{Hobbies:} Photography; Hiking; Travelling
\end{quote}

To ensure the accuracy of the Gemini-generated profiles, we conducted a human evaluation with statistical significance in mind. We randomly sampled 10\% of the generated profiles for manual review, ensuring a representative subset. Five human annotators independently assessed these profiles for correctness, clarity, and fabricated content. To measure inter-annotator agreement, we computed Fleiss’s kappa to quantify consistency among annotators. The verified profiles, designated as 'Gold Data,' served as a high-quality benchmark for validating and refining the generated profiles. This rigorous evaluation process ensured that the final dataset maintained a statistically validated standard of quality for training our models.

We studied how often each attribute appears in the dataset and saw that some attributes are much more common than others. This suggests that certain details are easier to find or are given more importance in the data.

\begin{figure}[!htb]
  \centering
  \includegraphics[width=\linewidth]{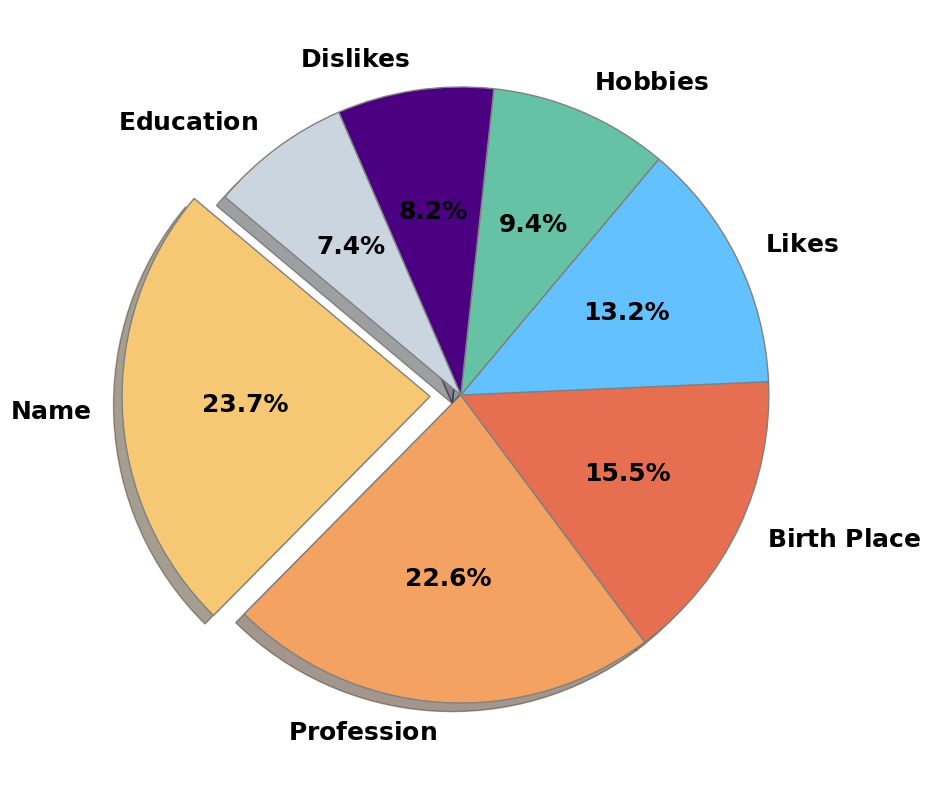}
  \caption{Distribution of Various Attributes in User Information}
  \label{fig:pi_chart}
\end{figure}

Figure~\ref{fig:pi_chart} shows that the most frequent attributes are:
\textbf{Name} (41{,}230 times), \textbf{Profession} (39{,}209), \textbf{Birth Place} (26{,}930), \textbf{Likes} (22{,}973), \textbf{Hobbies} (16{,}349), \textbf{Dislikes} (14{,}201), and \textbf{Education} (12{,}878).

Other attributes such as ``Born,'' ``Awards,'' ``Hobby,'' ``Birth date,'' ``Achievements,'' ``Occupation,'' and ``Nationality'' also appear often. 
Their frequency shows the dataset's focus and how it can be useful for various research tasks.

\begin{figure*}[!htb]
    \centering
    \begin{subfigure}{0.48\textwidth}
        \centering
        \includegraphics[width=\linewidth]{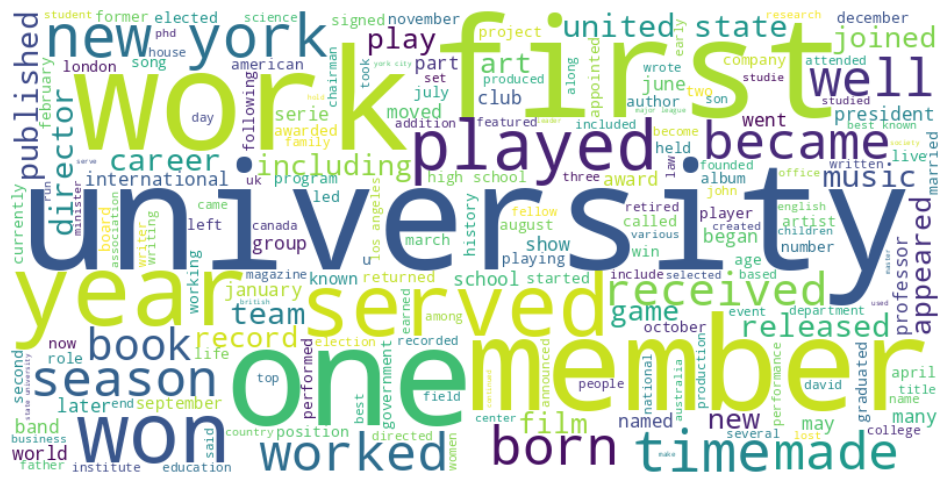}
        \caption{Word Cloud of Wiki People}
        \label{fig:word_cloud}
    \end{subfigure}
    \hfill
    \begin{subfigure}{0.48\textwidth}
        \centering
        \includegraphics[width=\linewidth]{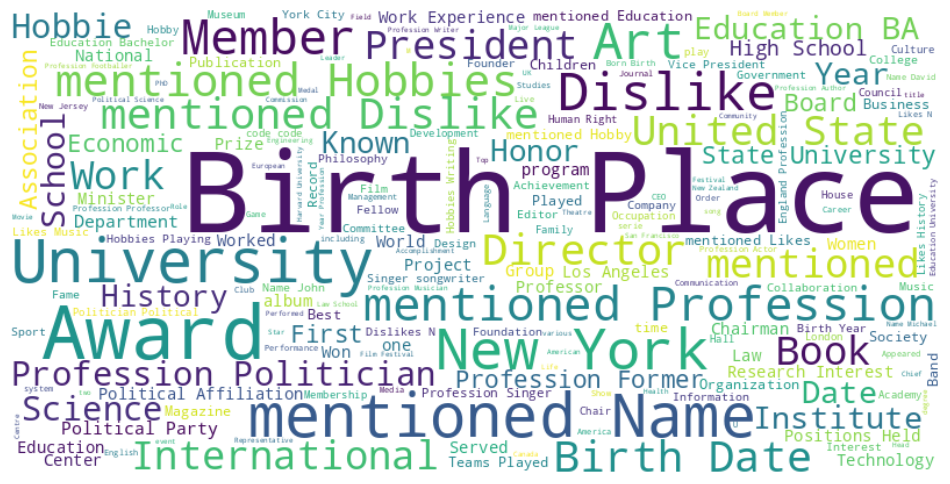}
        \caption{Word Cloud of Constructed User Profile}
        \label{fig:bar_chart}
    \end{subfigure}
    \caption{Visualization of the dataset: (a) Word cloud before constructing user profile dataset; (b) Word cloud after constructing user profile dataset.}
    \label{fig:dataset_visualization}
\end{figure*}

In Figure~\ref{fig:dataset_visualization}, we present the word cloud of ground truth and modified user profiles, illustrating the changes before and after updating user information.


\subsection{Updating User Profile Dataset}

We updated around 30\% of the user profiles to better reflect real-world changes, such as shifting interests or moving to a new location. 
Two types of modifications were made:

\textbf{(i). Removing Attributes or Elements:} 
We deleted certain items from existing attributes. 
For example, if the original profile listed 
\textbf{Hobbies:} Photography; Hiking; Travelling, 
we removed ``Hiking,'' resulting in 
\textbf{Hobbies:} Photography; Travelling.

\textbf{(ii). Introducing Contradictions or Changes:} 
We replaced some details to simulate evolving preferences. 
For instance, 
\textbf{Hobbies:} Photography; Hiking; Travelling 
could become 
\textbf{Hobbies:} Photography; Lounging. 
Here, the original set of hobbies (\emph{Photography, Hiking, Travelling}) represents the ground truth, while ``Lounging'' is a new, potentially contradictory element.

These modifications allow us to test how effectively our system can handle partial or altered information and restore profiles to match the original (ground truth) data. 
They also capture the dynamic nature of real-life scenarios where user interests and attributes can change over time.

\section{Problem Definition}


In this work, we tackle the challenge of building and updating user profiles using textual data. We work with a dataset \(D = \{(x_i, y_i)\}_{i=1}^{n}\), where each \(x_i\) represents a free-text biographical sketch of a user. This text can include personal details, interests, or other descriptive information, such as "John is a software developer who enjoys hiking and photography." The corresponding \(y_i\) is the user's profile, formatted as a structured list of attributes, like \{Profession: Software Developer, Hobbies: Hiking, Photography\}.

Our objective is to develop a model that accurately predicts \(y_i\) from \(x_i\). 

The conditional probability \(P(\bar{y}_i | x_i)\) represents the likelihood of generating a user profile \(\bar{y}_i\) given the input text \(x_i\). Here, \(\bar{y}_i\) is the predicted profile generated by the model, while \(y_i\) is the actual, ground truth profile. This probability reflects how well the model captures the correct attributes from the provided biographical context.

Our objectives are two-fold:
\paragraph{Profile Construction:} We train a model to predict a user profile \(\bar{y}_i\) from a biographical text \(x_i\). This involves estimating the conditional probability \(P(\bar{y}_i | x_i)\), capturing the relationship between the user's text and their profile attributes (see Figure \ref{fig:method}(a)).
\paragraph{Profile Updating:} After constructing an initial profile, it may require updates as new information \(x_i^u\) becomes available. For instance, if a user starts a new job or develops new interests, this new data must be integrated into the existing profile \(y_i\). The goal is to learn how to transition from the existing profile \(y_i\) to the updated profile \(y_i^u\) based on the new text \(x_i^u\) (see Figure \ref{fig:method}(b)).

\section{Model Construction}

Our method focuses on profile construction and updating using LLMs. Below, we detail the key components of our approach.

\subsection{Profile Construction}
We use a probabilistic framework to estimate the conditional probability \(P(\bar{y}_i | x_i)\), which represents the likelihood of generating a user profile \(\bar{y}_i\) from the input text \(x_i\). Here, \(x_i\) is treated as a sequential token input, typically a left-to-right sequence, aligning with common autoregressive training paradigms. 

We employ a pre-trained LLM due to its capability to handle sequential data and extract complex features. The model is defined as:
\begin{equation}
P(\bar{y}_i | x_i; \theta) = f_{\theta}(x_i)
\end{equation}
where \(f_{\theta}(\cdot)\) represents the LLM parameterized by \(\theta\).

We fine-tune the LLM using cross-entropy loss, which measures the difference between the predicted and ground truth profiles. The loss function is defined as:
\begin{equation}
L(\theta) = - \frac{1}{n} \sum_{i=1}^{n} \sum_{j=1}^{m} y_{i,j} \log P(\bar{y}_{i,j} | x_i; \theta)
\end{equation}
where \(y_{i,j}\) and \(\bar{y}_{i,j}\) are the true and predicted values for the \(j\)-th attribute of the \(i\)-th profile.

The Adam optimizer is used to update \(\theta\) iteratively, minimizing the loss \(L(\theta)\).

\subsection{Profile Updating}

Profile updating refines an existing user profile \(y_i\) with new biographical information \(x_i^u\). Here, \(x_i^u\) represents the most recent textual data, and \(y_i\) is the profile generated from previous data. We model this process by learning the conditional probability \(P(y_i^u | x_i^u, y_i; \zeta)\), where \(y_i^u\) is the updated profile, and \(\zeta\) represents the model parameters.

 We use a pre-trained LLM \(g_{\zeta}(\cdot)\) to model the updating process as follows:
\begin{equation}
P(\bar{y}_i^u | x_i^u, y_i; \zeta) = g_{\zeta}(x_i^u, y_i)
\end{equation}

The loss function for profile updating is the cross-entropy loss between the predicted and true updated profiles:
\begin{equation}
L_u(\zeta) = - \frac{1}{n} \sum_{i=1}^{n} \sum_{j=1}^{m} y_{i,j}^u \log P(\bar{y}_{i,j}^u | x_i^u, y_i; \zeta)
\end{equation}

\section{Experiments}
\subsection{Experimental Setup}

We conducted our experiments using two NVIDIA RTX H100 GPUs with 80GB of memory, employing the \texttt{PyTorch} framework. Model performance was evaluated using precision, recall, and \(F_1\) scores, supplemented by prompt-based evaluations with LLMs. We utilized the Transformers library from HuggingFace \cite{wolf2020transformers}, which provided a suite of tools and pretrained models tailored for natural language processing tasks. This facilitated efficient model training and evaluation across diverse datasets. Our experiments incorporated an ensemble of LLMs, including Bloom-7b, Llama2-7b, and Mistral-7b, to ensure comprehensive and robust results. In order to train the LLM model, we used Propulsion \cite{kowsher2024propulsion} PEFT instead of other PEFT methods due to its memory efficiency, faster training, and lower parameter requirements

\subsection{ Large Language Models (LLMs)}

For our analysis, we used an ensemble of LLMs to leverage their unique strengths and achieve robust results. This models included Bloom-7b, developed by BigScience, known for its strong performance on diverse NLP tasks \cite{le2023bloom}; Llama2-7b from Meta AI, offering versatile capabilities and fine-tuning options \cite{touvron2023llama}; Mistral-7b, an open-source model that outperforms larger counterparts on multiple benchmarks \cite{jiang2023mistral}; Falcon-7b, designed for a range of AI applications and trained on the REFINEDWEB dataset \cite{almazrouei2023falcon}; and Gemma-7b by Google, excelling in text summarization and code generation tasks \cite{team2024gemma}. Together, these models provided a comprehensive foundation for our user profiling tasks. 

\subsection{Evaluation Metrics}

To evaluate our approach, we employed two key metrics to assess the effectiveness of user profile construction and updating : the user-level \(F_1\) score and prompt-based evaluations via LLMs.

\subsubsection{User-level \texorpdfstring{$F_1$}{F1} Score}

Following the methodology of \citet{wen2023towards}, we calculated precision, recall, and the \(F_1\) score at the user level. For a user profile containing \(k\) attributes, the user-level \(F_1\) score is computed as follows:

\begin{equation}
\text{precision} = \frac{C(\text{Decoder}(\bar{y}_i), y_i)}{k - C^{\prime}(\text{Decoder}(\bar{y}_i), y_i)}
\label{eq:precision}
\end{equation}
\begin{equation}
\text{recall} = \frac{C^{\prime}(\text{Decoder}(\bar{y}_i), y_i)}{k}
\label{eq:recall}
\end{equation}
\begin{equation}
F_1 = 2 \cdot \frac{\text{precision} \cdot \text{recall}}{\text{precision} + \text{recall}}
\label{eq:f1}
\end{equation}

Here, \(C(\cdot)\) is the count of correctly predicted attributes, \(C^{\prime}(\cdot)\) represents the count of attributes with no prediction, and \(\text{Decoder}(\cdot)\) converts probabilistic outputs into the most likely token predictions.

\begin{table*}[!htb]
\centering
\scalebox{.75}{
\begin{tabular}{l|ccc|ccc||ccc|ccc}
\hline

\multicolumn{1}{c}{} & \multicolumn{6}{c||}{\textbf{Profile Construction}} & \multicolumn{6}{c}{\textbf{Updating Performance}} \\\hline
\multicolumn{1}{c}{} & \multicolumn{12}{c}{\textbf{(a) Zero-Shot Performance}} \\\hline
\multirow{2}{*}{Category} & \multicolumn{3}{c|}{F1 Score} & \multicolumn{3}{c||}{LLM Score} & \multicolumn{3}{c|}{F1 Score} & \multicolumn{3}{c}{LLM Score} \\ \cline{2-13} 
                          & Precision & Recall & \(F_1\) & Gemini & GPT-4 & Average & Precision & Recall & \(F_1\) & Gemini & GPT-4 & Average \\ \hline
Bloom-7b  & 75.12     & 68.54  & 71.65    & 78.43     & 76.25  & 77.34  & 74.83     & 69.12  & 71.85    & 80.13     & 79.75  & 79.94 \\
Llama2-7b    & 78.45     & 70.32  & 74.15    & 81.63     & 80.29  & 80.96  & 77.91     & 72.45  & 75.08    & 82.91     & 81.84  & 82.38 \\
Mistral-7b    & 79.65     & 72.18  & 75.71    & 83.24     & 84.01  & 83.62  & 79.93     & 73.58  & 76.54    & 84.51     & 85.12  & 84.82 \\
Falcon-7b   & 73.12     & 67.94  & 70.43    & 77.53     & 78.05  & 77.79  & 72.97     & 68.19  & 70.48    & 78.71     & 79.17  & 78.94 \\
Gemma-7b     & 76.34     & 70.25  & 73.15    & 80.65     & 79.98  & 80.32  & 75.81     & 71.14  & 73.38    & 81.94     & 80.58  & 81.26 \\\hline

\multicolumn{1}{c}{} & \multicolumn{12}{c}{\textbf{(b) Fine-Tuned Performance}} \\\hline
\multirow{2}{*}{Category} & \multicolumn{3}{c|}{F1 Score} & \multicolumn{3}{c||}{LLM Score} & \multicolumn{3}{c|}{F1 Score} & \multicolumn{3}{c}{LLM Score} \\ \cline{2-13} 
                          & Precision & Recall & \(F_1\) & Gemini & GPT-4 & Average & Precision & Recall & \(F_1\) & Gemini & GPT-4 & Average \\ \hline
Bloom-7b      & 94.74     & 83.74  & 90.42    & 95.13     & 93.95  & 94.54  & 94.83     & 85.64  & 92.68    & 96.63     & 96.75  & 96.69 \\
Llama2-7b     & 96.85     & 92.65  & 91.97    & 97.53     & 95.18  & 96.35  & 97.43     & 94.75  & 93.17    & 98.53     & 94.99  & 96.76 \\
Mistral-7b    & 97.17     & 93.65  & 93.84    & 98.04     & 99.01  & 98.52  & 97.93     & 95.02  & 95.08    & 97.91     & 99.18  & 98.54 \\
Falcon-7b     & 92.88     & 89.92  & 90.47    & 95.43     & 96.05  & 95.74  & 93.97     & 91.19  & 93.13    & 97.01     & 97.17  & 97.09 \\
Gemma-7b      & 95.63     & 92.87  & 91.15    & 97.95     & 96.89  & 97.42  & 96.31     & 93.48  & 94.53    & 98.94     & 96.18  & 97.56 \\\hline
\end{tabular}}
\caption{Performance comparison of user profile construction and updating. Part (a) shows zero-shot results, and part (b) presents fine-tuned results, highlighting precision, recall, \(F_1\) scores, and LLM evaluation scores from Gemini and GPT-4.}
\label{tab:up}
\end{table*}

\subsubsection{Prompt-based LLM Evaluation}

We utilized prompt-based assessments with advanced LLMs, specifically GPT-4 and Gemini Pro, for a detailed evaluation. This involved crafting prompts that framed the task of assessing the accuracy of constructed user profiles. An example prompt is:

\begin{quote}
\textit{``Given the context of user information and a constructed user profile, evaluate the profile by assigning a score from 0 to 1 based on its correctness. A score of 1 indicates complete accuracy, while 0 indicates total inaccuracy. Consider each attribute in the profile, adjusting the overall score based on any inaccuracies.''}
\end{quote}

This method provided a quantitative score for each user profile, and we averaged these scores to gauge overall model performance.

\subsection{Hyperparameters Setting and Model Training}

For model training and hyperparameter tuning, we adopted a focused approach to identify optimal parameters. We selected a learning rate of \(1e^{-4}\) using logarithmic scaling and set the batch size to 16 to balance efficiency and accuracy. Training involved a weight decay of 0.1, a dropout rate of 0.2, and an attention dropout rate of 0.1 over a single epoch.

\subsection{Result Analysis}

Our evaluation includes two phases: profile construction and profile updating, as shown in Table \ref{tab:up}. In both phases, we examined performance using precision, recall, \(F_1\) scores, and LLM evaluation scores from Gemini and GPT-4.

\subsubsection{Profile Construction Analysis}
In the profile construction phase (Table \ref{tab:up}, left), we observed a significant improvement in performance after fine-tuning compared to zero-shot settings. 

In the zero-shot scenario, \textbf{Mistral-7b} achieved an \(F_1\) score of 75.71, followed by \textbf{Llama2-7b} at 74.15. While these models exhibited decent zero-shot performance, their precision and recall were notably lower compared to their fine-tuned counterparts. For instance, \textbf{Mistral-7b} improved from a precision of 79.65 (zero-shot) to 97.17 (fine-tuned), and recall increased from 72.18 to 93.65.

After fine-tuning, \textbf{Mistral-7b} led with an \(F_1\) score of 93.84, followed by \textbf{Llama2-7b} at 91.97. Both models exhibited high precision (\textbf{Llama2-7b}: 96.85, \textbf{Mistral-7b}: 97.17) and strong recall (\textbf{Mistral-7b}: 93.65, \textbf{Llama2-7b}: 92.65), demonstrating their enhanced capability in accurately identifying user attributes. LLM scores further highlighted \textbf{Mistral-7b}'s robustness, particularly in the \textbf{GPT-4} evaluation (99.01).

\subsubsection{Profile Updating Analysis}
In the profile updating phase (Table \ref{tab:up}, right), similar trends were observed. In the zero-shot setting, \textbf{Mistral-7b} achieved an \(F_1\) score of 76.54, with precision and recall at 79.93 and 73.58, respectively. These scores highlight the model's baseline capability but also indicate room for improvement.

Post fine-tuning, \textbf{Mistral-7b} achieved the highest \(F_1\) score of 95.08, with precision at 97.93 and recall at 95.02, showcasing a substantial improvement over the zero-shot results. \textbf{Gemma-7b} also showed strong performance, improving from a zero-shot \(F_1\) score of 73.38 to 94.53 after fine-tuning. Precision for \textbf{Llama2-7b} increased from 77.91 (zero-shot) to 97.43 (fine-tuned), and \textbf{Falcon-7b} showed improvements across all metrics as well.

LLM evaluations reaffirmed \textbf{Mistral-7b}'s strength, especially in the GPT-4 assessment, where its score increased from 85.12 (zero-shot) to 99.18 (fine-tuned). \textbf{Gemma-7b} also displayed consistent performance improvements across both Gemini and GPT-4 evaluations.

\section{Conclusion}

In this work, we have introduced a robust methodology for the construction and updating of user profiles using large language models (LLMs). By leveraging the probabilistic framework, we were able to model the conditional distribution of user profiles based on input text sequences, leading to highly accurate and contextually relevant profile generation. The extension of this approach to profile updating further demonstrated the adaptability of our method, allowing existing profiles to be refined and enhanced as new information becomes available.

Our experimental results underscore the effectiveness of the proposed approach, with models like Mistral-7b and Llama2-7b achieving strong performance across both profile construction and updating tasks. The use of LLMs not only improved the precision and recall of the profiles generated but also ensured that these profiles were in alignment with advanced evaluative frameworks, as evidenced by the high scores in LLM assessments.

\bibliography{acl_latex}

\appendix

\section{Examples of User Profile Construction Using LLM}
\textbf{Example-1:}

\textit{\textcolor{red}{\textbf{User Information:}} Meredith ochs is an american radio commentator dj writer and musicianochs is a commentator for national public radios all things considered and has appeared on talk of the nation she is also a frequent guest on all songs consideredochs is a talk show host at sirius xm satellite radio the program freewheelin with meredith ochs chris t airs on the road dog trucking channel sirius xm 106 weekdays 11 am 2 pm et replays weeknights 811 pm et saturday and sunday 11 am 2 pm et ochs also hosts a music show on the outlaw country channel sirius xm 60 weeknights 8 pm midnight et ochs was a contributing editor and writer at guitar world magazine where she began working in 1996 her writing has appeared in numerous other publications including entertainment weekly rolling stone saloncom the boston phoenix the houston chronicle and new york magazine her photographs have appeared in guitar world experience hendrix and other publicationsochs plays guitar sings and writes songs for the damn lovelys their debut trouble creek was released on philadelphia indie label dren records in 2004 and earned them an appearance on mountain stage the internationally syndicated pri radio program the following year she also plays bass and sings with new orleansborn rbsoul singer jesse thomas known as young jesse when he worked with huey piano smith earl king and other new orleans performersfor more than a decade ochs hosted a weekly show trash twang and thunder on freeform radio station wfmu the show focused on american roots music country blues soul rb and bluegrass and often featured live performances by artists including son volts jay farrar country singer pam tillis and allman brothers alumnus derek trucks ochs later cohosted with and filled in for fmu alum vin scelsa at wfuv where she worked parttime as a dj ochs also worked at wnyc and wxrk krock in new york city as well as the abc radio network}

\textit{\textcolor{blue}{\textbf{Output:}}}

\textbf{Name:} Meredith Ochs

\textbf{Profession:} Radio commentator, DJ, writer, musician

\textbf{Likes:} American roots music, country blues, soul, R\&B, bluegrass

\textbf{Hobbies:} Playing guitar, singing, writing songs

\textbf{Work:}
\begin{itemize}
    \item Commentator for National Public Radio's All Things Considered
    \item Talk show host at Sirius XM Satellite Radio
    \item Contributing editor and writer at Guitar World magazine
    \item DJ at WFUV and WFMU
\end{itemize}

\textbf{Additional Information:}
\begin{itemize}
    \item Plays guitar, sings, and writes songs for the Damn Lovelys
    \item Plays bass and sings with New Orleans-born R\&B/soul singer Jesse Thomas
    \item Has appeared on Mountain Stage and Talk of the Nation
\end{itemize}

\textbf{Example-2:}

\textit{\textcolor{red}{\textbf{User Information:}} 
David chernushenko born june 1963 in calgary alberta is a politician professional speaker sustainability consultant and documentary filmmaker in ontario canada he was elected to ottawa city council in the 2010 municipal election and is running for reelection in 2014 he is the former senior deputy to the leader of the green party of canada and a former leadership contestant for that partya graduate of queens university political science and alumnus of cambridge university international relations chernushenko has worked for the canadian international development agency cida the canadian department of foreign affairs and international trade dfait and the united nations environment programme he has served on committees and boards of local housing and environment groups schools and health advisory bodieschernushenko is a green building professional accredited by the leed leadership in energy and environmental design certification program as owner of the consulting firm green gold inc since 1998 he has advised public private and nonprofit organizations on adopting more sustainable and socially responsible practicesfrom 1998 to 2004 chernushenko served on the international olympic committees commission on sport and the environment he has written several books on sustainable management practices including sustainable sport management unep 2001 and greening our games running sports events facilities that wont cost the earth centurion 1994 and the electronic publication greening campuses and their communities iisdacccunep 1996 in 2001 he cofounded clean air champions a national charity that engages athletes in raising awareness about air pollution climate change and the benefits of physical activity in canada he is a director of the sustainable ottawa energy cooperativehe was the green party candidate for ottawa centre in the 2004 federal election he finished fourth with 4730 votes 8 receiving more votes than any other green candidate in ottawa he ran again in ottawa centre in the 2006 federal election and again came fourth losing to paul dewar from the new democratic party ndp chernushenko received 6766 votes 102 the highest vote count of any green party candidate in canada the 2006 election he passed the 10 threshold thus becoming eligible for partial government reimbursement of campaign expenditures chernushenko was endorsed by the ottawa citizen newspaper in both the 2004 and 2006 elections he also ran as the green party candidate in ottawa south in the 2003 ontario general electionon november 10 2006 prime minister stephen harper appointed chernushenko to canadas national round table on the environment and the economy a nonpartisan panel that advises the federal government on environmental policy and works to promote the integrated pursuit of excellence in environmental social and economic practices in canadas public private and civil society sectors chernushenko served as a member on the nrtee from 2006 to 2009 and served as vicechair in 20082009chernushenko resigned as deputy leader of the green party in july 2007 in order to devote more time to his international consulting business and the nrtee and to make documentary films since then he has launched the living lightly multimedia project and produced three documentaries titled be the change 2008 powerful energy for everyone 2010 and bike city great city 2013chernushenko ran as city councillor for capital ward in the 2010 ottawa municipal election he won with 4134 of the vote he is running for reelection in 2014
}

\textit{\textcolor{blue}{\textbf{Output:}}}

\textbf{Name:} Timothy Gruzca

\textbf{Born:} 1 July 1976, Melbourne, Australia

\textbf{Profession:} Cameraman, Documentary Film Maker

\textbf{Based In:} Paris, France

\textbf{Likes:}
\begin{itemize}
    \item Working in conflict zones
    \item Telling stories of people affected by war
    \item Documenting humanitarian crises
\end{itemize}

\textbf{Dislikes:}
\begin{itemize}
    \item War
    \item Injustice
    \item Suffering
\end{itemize}

\textbf{Hobbies:}
\begin{itemize}
    \item Photography
    \item Hiking
    \item Travelling
\end{itemize}

\section{Examples of User Profile Updating Using LLM}
\textbf{Example-1:}

 \textit{\textcolor{red}{\textbf{User Information:}}
Jonathan hoefler born august 22 1970 is an american typeface designer hoefler pronounced heffler founded the hoefler type foundry in 1989 a type foundry in new york in 1999 hoefler began working with type designer tobias frerejones and from 20052014 the company operated under the name hoefler frerejones until their public splithoefler has designed original typefaces for rolling stone magazine harpers bazaar the new york times magazine sports illustrated and esquire and several institutional clients including the solomon r guggenheim museum and alternative band they might be giants perhaps his bestknown work is the hoefler text family of typefaces designed for apple computer and now appearing as part of the macintosh operating system he also designed the current wordmark of the church of the latterday saintsin 1995 hoefler was named one of the forty most influential designers in america by id magazine and in 2002 the association typographique internationale atypi presented him with its most prestigious award the prix charles peignot for outstanding contributions to type design hoefler and frerejones have been profiled in the new york times time magazine and esquire magazine and appearances on national public radio and cbs sunday morninghoeflers work is part of the cooperhewitt national design museums permanent collection.}

\textit{\textcolor{red}{\textbf{User Profile Memory:}}}

\textbf{Name:} Jonathan

\textbf{Birth Date:} October 18, 1977

\textbf{Company:} Hoefler Type Foundry

\textbf{Awards:} Lifetime Achievement Award

\textbf{Media Appearances:}
\begin{itemize}
    \item Esquire Magazine
\end{itemize}

\textbf{Collections:}
\begin{itemize}
    \item The National Design Museum
\end{itemize}

\textbf{Likes:} Time Magazine

\textbf{Dislikes:} Cooper-Hewitt National Design Museum

\textcolor{blue}{\textbf{Output:}}

\textbf{Name:} Jonathan Hoefler

\textbf{Birth Date:} August 22, 1970

\textbf{Profession:} Typeface Designer

\textbf{Company:} Hoefler Type Foundry

\textbf{Awards:} Prix Charles Peignot

\textbf{Design Work:}
\begin{itemize}
    \item Hoefler Text Family of Typefaces
    \item Wordmark of the Church of the Latter-day Saints
\end{itemize}

\textbf{Media Appearances:}
\begin{itemize}
    \item The New York Times
    \item Time Magazine
    \item Esquire Magazine
    \item National Public Radio
    \item CBS Sunday Morning
\end{itemize}

\textbf{Collections:}
\begin{itemize}
    \item Cooper-Hewitt National Design Museum
\end{itemize}

\begin{table*}[htp!]
\centering
\resizebox{\textwidth}{!}{%
\begin{tabular}{|l|l|p{10cm}|}
\hline
\rowcolor{gray!20} 
\textbf{URI} & \textbf{Name} & \textbf{Summary} \\ \hline
\rowcolor[HTML]{D0E7F7} <http://dbpedia.org/resource/Digby\_Morrell> & Digby Morrell & Former Australian rules footballer who played for the Kangaroos and Carlton in the AFL. Leading goalkicker for West Perth in 2000, traded to Carlton in 2003, and later coached Strathmore and West Coburg football clubs. Currently teaches physical education at Parade College in Melbourne. \\ \hline
\rowcolor[HTML]{F3DBD7} <http://dbpedia.org/resource/Alfred\_J.\_Lewy> & Alfred J. Lewy & Professor and researcher in psychiatry at OHSU, specializing in chronobiologic, sleep, and mood disorders. Known for his work on melatonin and bright light therapy for circadian rhythm disorders. \\ \hline
\rowcolor[HTML]{D8DBF5} <http://dbpedia.org/resource/Harpdog\_Brown> & Harpdog Brown & Canadian blues singer and harmonica player active since 1982. Known for albums like "Home is Where the Harp Is" and "What It Is." Awarded the Maple Blues Award for Best Harmonica Player in Canada (2014). \\ \hline
\rowcolor[HTML]{D0E7F7} <http://dbpedia.org/resource/Franz\_Rottensteiner> & Franz Rottensteiner & Austrian publisher and critic in science fiction and fantasy. Edited the journal "Quarber Merkur" and translated works by Stanislaw Lem and Philip K. Dick. Known for his critical views on American science fiction. \\ \hline
\rowcolor[HTML]{F3DBD7} <http://dbpedia.org/resource/G-Enka> & G-Enka & Estonian rapper and record producer, known for his work with the band Toe Tag and collaborations with ArhM. Founded Legendaarne Records and performed as a warm-up act for 50 Cent in 2007. \\ \hline
\rowcolor[HTML]{D8DBF5} <http://dbpedia.org/resource/Sam\_Henderson> & Sam Henderson & American cartoonist and writer, creator of "The Magic Whistle" and "Scene but Not Heard." Nominated for Harvey Awards and an Emmy for his work on "SpongeBob SquarePants." \\ \hline
\rowcolor[HTML]{D0E7F7} <http://dbpedia.org/resource/Aaron\_LaCrate> & Aaron LaCrate & American music producer, DJ, and fashion designer. Known for popularizing Baltimore club music and founding MilkCrate Records. Collaborated with HBO for "The Wire" and produced music for Dizzee Rascal and Madonna. \\ \hline
\rowcolor[HTML]{F3DBD7} <http://dbpedia.org/resource/Trevor\_Ferguson> & Trevor Ferguson & Canadian novelist and playwright, also known as John Farrow. Author of "City of Ice" and "River City." His works have been highly acclaimed in France and adapted into films and miniseries. \\ \hline
\rowcolor[HTML]{D8DBF5} <http://dbpedia.org/resource/Grant\_Nelson> & Grant Nelson & English DJ and record producer, known as a pioneer of UK garage. Produced under aliases like Wishdokta and Bump Flex. Remixed artists such as Kelis, Jamiroquai, and Faithless. \\ \hline
\rowcolor[HTML]{D0E7F7} <http://dbpedia.org/resource/Cathy\_Caruth> & Cathy Caruth & Professor at Cornell University, specializing in trauma theory. Author of "Unclaimed Experience" and editor of "Trauma: Explorations in Memory." Known for her innovative work on trauma and narrative. \\ \hline
\end{tabular}%
}
\caption{Summary Wiki People Dataset} \label{tab:overloaded_summary}
\end{table*}

\end{document}